\documentclass[11pt,letterpaper]{article}
\usepackage{emnlp2016}
\usepackage{times}
\usepackage{latexsym}
\usepackage{graphicx}
\usepackage{epstopdf}
\usepackage{longtable}
\usepackage{tabularx}
\usepackage{amsfonts}
\usepackage{amsmath}
\usepackage{enumerate}
\usepackage{mathtools}
\usepackage{tikz}
\usepackage{float}
\usepackage{fancyvrb}
\usepackage{pdflscape}
\usepackage{multirow}
\usepackage{enumitem}
\usepackage{booktabs}
\usepackage{amsmath}
\usepackage{algorithm}
\usepackage{algpseudocode}
\usepackage{xr}
\usepackage{subcaption}

\emnlpfinalcopy

\def\emnlppaperid{858}

\DeclareMathOperator*{\argmax}{arg\,max}

\newcommand{\Break}{\State \textbf{break} }

\title{Online Segment to Segment Neural Transduction}

  \author{Lei Yu$^1$,  Jan Buys$^1$ \and Phil Blunsom$^{1,2}$ \\
  	        $^1$University of Oxford \\  $^2$DeepMind \\
  {\tt \{lei.yu, jan.buys, phil.blunsom\}@cs.ox.ac.uk}}

\date{}

\begin{document}

\maketitle

\begin{abstract}
We introduce an online neural sequence to sequence model that learns to alternate between encoding and decoding segments of the input as it is read. By independently tracking the encoding and decoding representations our algorithm permits exact polynomial marginalization of the latent segmentation during training, and during decoding beam search is employed to find the best alignment path together with the predicted output sequence. Our model tackles the bottleneck of vanilla encoder-decoders that have to read and memorize the entire input sequence in their fixed-length hidden states before producing any output. It is different from previous attentive models in that, instead of treating the attention weights as output of a deterministic function, our model assigns attention weights to a sequential latent variable which can be marginalized out and permits online generation. 
Experiments on abstractive sentence summarization and morphological inflection show significant performance gains over the baseline encoder-decoders. 
\end{abstract}

\section{Introduction}
The problem of mapping from one sequence to another is an importance challenge of natural language processing. Common applications include machine translation and abstractive sentence summarisation.
Traditionally this type of problem has been tackled by a combination of hand-crafted features, alignment models, segmentation heuristics, and language models, all of which are tuned separately. 

The recently introduced encoder-decoder paradigm has proved very successful for machine translation, where an input sequence is encoded into a fixed-length vector and an output sequence is then decoded from said vector \cite{kalchbrenner2013recurrent,sutskever2014sequence,DBLP:conf/emnlp/ChoMGBBSB14}. 
This architecture is appealing, as it makes it possible to tackle the problem of sequence-to-sequence mapping by training a large neural network in an end-to-end fashion. However it is difficult for a fixed-length vector to memorize all the necessary information of an input sequence, especially for long sequences. Often a very large encoding needs to be employed in order to capture the longest sequences, which invariably wastes capacity and computation for short sequences. 
While the attention mechanism of \newcite{DBLP:journals/corr/BahdanauCB14} goes some way to address this issue, it still requires the full input to be seen before any output can be produced.

In this paper we propose an architecture to tackle the limitations of the vanilla encoder-decoder model, a segment to segment neural transduction model (SSNT) that learns to generate and align simultaneously.
Our model is inspired by the HMM word alignment model proposed for statistical machine translation \cite{vogel1996hmm,tillmann1997dp}; we impose a monotone restriction on the alignments but incorporate recurrent dependencies on the input which enable rich locally non-monotone alignments to be captured. This is similar to the sequence transduction model of \newcite{graves2012sequence}, but we propose alignment distributions which are parameterised separately, making the model more flexible and allowing online inference. 

Our model introduces a latent segmentation which determines correspondences between tokens of the input sequence and those of the output sequence. The aligned hidden states of the encoder and decoder are used to predict the next output token and to calculate the transition probability of the alignment. We carefully design the input and output RNNs such that they independently update their respective hidden states. This enables us to derive an exact dynamic programme to marginalize out the hidden segmentation during training and an efficient beam search to generate online the best alignment path together with the output sequence during decoding. Unlike previous recurrent segmentation models that only capture dependencies in the input \cite{graves2006connectionist,kong2015segmental}, our segmentation model is able to capture unbounded dependencies in both the input and output sequences while still permitting polynomial inference. 

While attentive models treat the attention weights as output of a deterministic function, our model assigns attention weights to a sequential latent variable which can be marginalized out.
Our model is general and could be incorporated into any RNN-based encoder-decoder architecture, such as Neural Turing Machines \cite{graves2014neural}, memory networks \cite{weston2014memory,kumar2015ask} or stack-based networks \cite{grefenstette2015learning}, enabling such models to process data online.

We conduct experiments on two different transduction tasks, abstractive sentence summarisation (sequence to sequence mapping at word level) and morphological inflection generation (sequence to sequence mapping at character level). We evaluate our proposed algorithms in both the online setting, where the input is encoded with a unidirectional LSTM, and where the whole input is available such that it can be encoded with a bidirectional network.
The experimental results demonstrate the effectiveness of SSNT --- it consistently output performs the baseline encoder-decoder approach while requiring significantly smaller hidden layers, thus showing that the segmentation model is able to learn to break one large transduction task into a series of smaller encodings and decodings.
When bidirectional encodings are used the segmentation model outperforms an attention-based benchmark.
Qualitative analysis shows that the alignments found by our model are highly intuitive and demonstrates that the model learns to read ahead the required number of tokens before producing output.

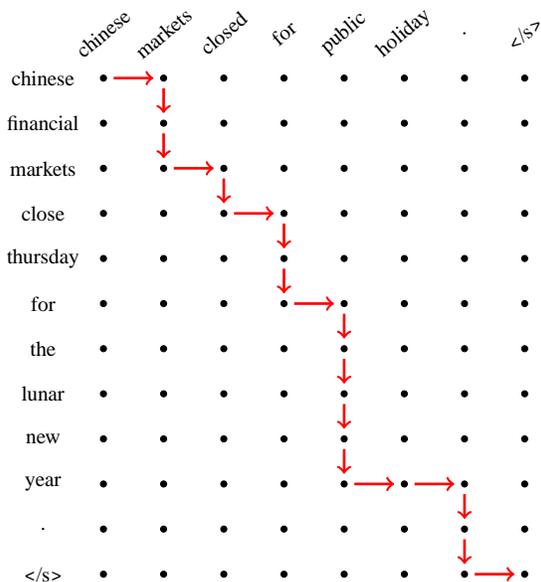
\begin{figure}
	\begin{tikzpicture}
	\def \xscale {0.8}
	\def \yscale {0.6}
	
	\foreach \x in {0,...,7}
	\foreach \y in {0,...,11} 
	{\draw [fill]  (\xscale*\x, \y*\yscale) circle [radius=0.04] ;
		\node   (\x\y) at (\xscale*\x,\y*\yscale) {};} 
	
	\foreach \c [count=\x from 0] in {
		{\texttt{<}/s\texttt{>}},{.},{year},{new},{lunar},{the},{for},{thursday},{close},{markets}, {financial},{chinese}} 
	{
		\node at (-\xscale,\x*\yscale) {\scriptsize \c};	
	}

	\foreach \c [count=\x from 0] in {
		{chinese},{markets},{closed},{for}, {public},{holiday}, {.},{\texttt{<}/s\texttt{>}}} 
	{
		\node [rotate=40] at (\xscale*\x,12*\yscale) {\scriptsize \c};	
	} 
	
	\draw [->,red,line width=1] 
	(011) edge (111) 
	(111) edge (110)
	(110) edge (19)
	(19) edge (29)
	(29) edge (28)
	(28) edge (38)
	(38) edge (37)
	(37) edge (36)
	(36) edge (46)
	(46) edge (45)
	(45) edge (44)
	(44) edge (43)
	(43) edge (42)
	(42) edge (52)
	(52) edge (62)
	(62) edge (61)
	(61) edge (60)
	(60) edge (70);
	\end{tikzpicture}
	\caption{Example output of our recurrent segmentation model on the task of abstractive sentence summarisation. The path highlighted is the alignment found by the model during decoding. 
		}
	\label{example_graph}
\end{figure}

\section{Model}
Let $\boldsymbol x_1^I$ be the input sequence of length $I$ and $\boldsymbol{y}_1^J$ the output sequence of length $J$. Let $y_j$ denote the $j$-th token of $\boldsymbol{y}$. Our goal is to model the conditional distribution
\begin{equation}
p(\boldsymbol{y}|\boldsymbol{x}) = \prod_{j=1}^{J} p(y_j| \boldsymbol{y}_1^{j-1}, \boldsymbol{x}).
\end{equation}
We introduce a hidden alignment sequence $\boldsymbol{a}_1^J$ where each $a_j = i$ corresponds to an input position $i \in \{1, \dots, I\}$ that we want to focus on when generating $y_j$. Then $p(\boldsymbol{y}|\boldsymbol{x})$ is calculated by marginalizing over all the hidden alignments,

\begin{eqnarray}
\label{con_prob}
&p(\boldsymbol{y}|\boldsymbol{x})  = \sum_{\boldsymbol{a}} p(\boldsymbol{y}, \boldsymbol{a} | \boldsymbol{x}) \\
\nonumber \approx & \sum_{\boldsymbol{a}}  \prod_{j=1}^{J} \underbrace{p(a_j | a_{j-1}, \boldsymbol{y}_1^{j-1}, \boldsymbol{x})}_{\text{transition probability}} \cdot \\
\nonumber & \ \ \ \ \ \ \ \ \ \ \ \ \ \ \ \ \ \ \ \ \ \ \ \ \ \ \ \ \ \ \ \ \ \ \ \ \ \ \underbrace{p(y_j | \boldsymbol{y}_1^{j-1}, a_j, \boldsymbol{x}).}_{\text{word prediction}}
\end{eqnarray}

Figure \ref{example_graph} illustrates the model graphically. Each path from the top left node to the right-most column in the graph corresponds to an alignment.
We constrain the alignments to be monotone, i.e. only forward and downward transitions are permitted at each point in the grid.
This constraint enables the model to learn to perform online generation. 
Additionally, the model learns to align input and output segments, which means that it can learn local reorderings by memorizing phrases. 
Another possible constraint on the alignments would be to ensure that the entire input sequence is consumed before last output word is emitted, i.e. all valid alignment paths have to end in the bottom right corner of the grid. However, we do not enforce this constraint in our setup.

The probability contributed by an alignment is obtained by accumulating the probability of word predictions at each point on the path and the transition probability between points.
The transition probabilities and the word output probabilities are modeled by neural networks, which are described in detail in the following sub-sections.

\subsection{Probabilities of Output Word Predictions}
The input sentence $\boldsymbol{x}$ is encoded with a Recurrent Neural Network (RNN), in particular an LSTM \cite{hochreiter1997long}. The encoder can either be a unidirectional or bidirectional LSTM. If a unidirectional encoder is used the model is able to read input and generate output symbols online. The hidden state vectors are computed as 
\begin{align}
\mathbf{h}_i^\rightarrow &= \text{RNN}(\mathbf{h}_{i-1}^\rightarrow, v^{(e)}(x_i)), \\
\mathbf{h}_i^\leftarrow &= \text{RNN}(\mathbf{h}_{i+1}^\leftarrow ,v^{(e)}(x_i)) ,
\end{align}
where $v^{(e)}(x_i)$ denotes the vector representation of the token $x$, and $\mathbf{h}_i^\rightarrow$ and $\mathbf{h}_i^\leftarrow$ are the forward and backward hidden states, respectively. For a bidirectional encoder, they are concatenated as $\mathbf{h}_i = [\mathbf{h}_i^\rightarrow; \mathbf{h}_i^\leftarrow]$; and for unidirectional encoder $\mathbf{h}_i = \mathbf{h}_i^\rightarrow$. The hidden state $\mathbf{s}_j$ of the RNN for the output sequence $\boldsymbol{y}$ is computed as 
\begin{equation}
\mathbf{s}_j = \text{RNN}(\mathbf{s}_{j-1}, v^{(d)}(y_{j-1})), 
\end{equation}
where $v^{(d)}(y_{j-1})$ is the encoded vector of the previously generated output word $y_{j-1}$. That is, $\mathbf{s}_j$ encodes $\boldsymbol{y}_1^{j-1}$.

To calculate the probability of the next word, we concatenate the aligned hidden state vectors $\mathbf{s}_j$ and $\mathbf{h}_{a_j}$ and feed the result into a softmax layer,
 \begin{equation}
 \begin{split}
 &\ p(y_j = l | \boldsymbol{y}_1^{j-1}, a_j, \boldsymbol{x}) \\
 = &\ p(y_j = l | \mathbf{h}_{a_j}, \mathbf{s}_j)\\
 = &\ \text{softmax} (\mathbf{W}_w[\mathbf{h}_{a_j};\mathbf{s}_j] + \mathbf{b}_w)_l.
 \end{split}
 \end{equation}
The word output distribution in \newcite{graves2012sequence} is parameterised in similar way.

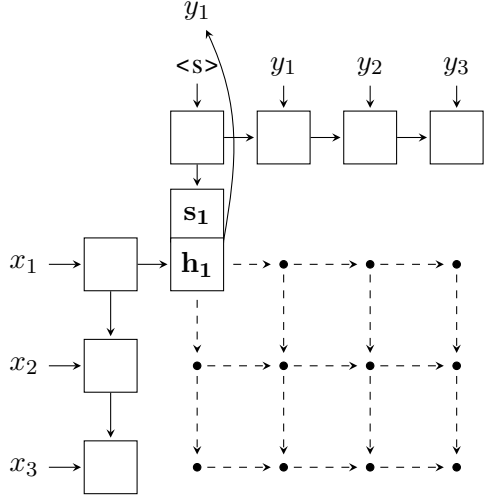
\begin{figure}
	\begin{tikzpicture}[>=stealth,shorten >=1pt,auto]
	\def \xscale {1.15}
	\def \yscale {1.35}
	\def \rectangleSize {20}
	
	\tikzstyle{rectangle_style}=[draw,fill=white,inner sep=0pt,minimum width=\rectangleSize,minimum height=\rectangleSize]
	
	\node (x3) at (1*\xscale,0)  {$x_3$}; 
	\node (x2) at (1*\xscale,1*\yscale)  {$x_2$}; 
	\node (x1) at (1*\xscale,2*\yscale)  {$x_1$};

	\foreach \x in {0,...,3}
	\foreach \y in {0,...,2} 
	{\draw [fill]  (\xscale*3+\xscale*\x, \y*\yscale) circle [radius=0.05] ;
		\node   (\x\y) at (\xscale*3+\xscale*\x,\y*\yscale) {};}
	
	\node (x12)[rectangle_style] at (\xscale*2,2*\yscale) {};
	
	\node (x13up)[rectangle_style] at (\xscale*3,2.48*\yscale) {$\bf s_1$};
	\node (x13below)[rectangle_style] at (\xscale*3,2*\yscale) {$\bf h_1$};
	
	\node (x22)[rectangle_style] at (\xscale*2,\yscale) {};
	
	\node (x32)[rectangle_style] at (\xscale*2,0) {};
	
	\node (y1) at (\xscale*3,\yscale*3.95) {\texttt{<}s\texttt{>}};
	\node (y11)[rectangle_style] at (\xscale*3,\yscale*3.25) {};
	
	\node (y2) at (\xscale*4,\yscale*3.95) {$y_1$};
	\node (y21)[rectangle_style] at (\xscale*4,\yscale*3.25) {};
	
	\node (y3) at (\xscale*5,\yscale*3.95) {$y_2$};
	\node (y31)[rectangle_style] at (\xscale*5,\yscale*3.25) {};
	
	\node (y4) at (\xscale*6,\yscale*3.95) {$y_3$};
	\node (y41)[rectangle_style] at (\xscale*6,\yscale*3.25) {};

	\draw [->] 
	(x1) edge (x12)
	(x12) edge (x13below.west)
	(x2) edge (x22)
	(x12) edge (x22)
	(x3) edge (x32)
	(x22) edge (x32)
	;
	
	\draw [->] 
	(y11) edge (y21)
	(y11) edge (x13up)
	(y21) edge (y31)
	(y31) edge (y41)
	(y1) edge (y11)
	(y2) edge (y21)
	(y3) edge (y31)
	(y4) edge (y41)
	;
	
	\node (between) at (\xscale*3.28,\yscale*2.11) {};
	\draw [<-]
	node at (\xscale*3,\yscale*4.5) {$y_1$}  edge[in=80,out=295] (between);

	\draw [->,dashed]
	node at (\xscale*3.3,\yscale*2) {} edge (12)
	node at (\xscale*3,\yscale*1.75) {} edge (01)
	(12) edge (22)
	(12) edge (11)
	(22) edge (32)
	(22) edge (21)
	(32) edge (31)
	(11) edge (21)
	(11) edge (10)
	(21) edge (31)
	(21) edge (20)
	(31) edge (30)
	(01) edge (00)
	(01) edge (11)
	(00) edge (10)
	(10) edge (20)
	(20) edge (30)
	;
	\end{tikzpicture}
	\caption{The structure of our model. $(x_1, x_2, x_3)$ and $(y_1, y_2, y_3)$ denote the input and output sequences, respectively. The points, e.g. $(i,j)$, in the grid represent an alignment between $x_i$ and $y_j$. For each column $j$, the concatenation of the hidden states $[\mathbf{h}_i, \mathbf{s}_j]$ is used to predict $y_{j}$.}
	\label{model_structure}
\end{figure}

Figure \ref{model_structure} illustrates the model structure. 
Note that the hidden states of the input and output decoders are kept independent to permit tractable inference, while the output distributions are conditionally dependent on both.
 
\subsection{Transition Probabilities}

As the alignments are constrained to be monotone, we can treat the transition from timestep $j$ to $j+1$ as a sequence of \texttt{shift} and \texttt{emit} operations. Specifically, at each input position, a decision of \texttt{shift} or \texttt{emit} is made by the model; if the operation is \texttt{emit} then the next output word is generated; otherwise, the model will \texttt{shift} to the next input word.
While the multinomial distribution is an alternative for parameterising alignments, the shift/emit parameterisation does not place an upper limit on the jump size, as a multinomial distribution would, and biases the model towards shorter jump sizes, which a multinomial model would have to learn.

We describe two methods for modelling the alignment transition probability. The first approach is independent of the input or output words. To parameterise the alignment distribution in terms of shift and emit operations we use a geometric distribution,
\begin{equation}
p(a_j|a_{j-1}) = (1-e)^{a_j - a_{j-1}} e,
\end{equation}
where $e$ is the emission probability. This transition probability only has one parameter $e$, which can be estimated directly by maximum likelihood as
\begin{equation}
e = \frac{\sum_n J_n}{\sum_n I_n + \sum_n J_n},
\end{equation}
where $I_n$ and $J_n$ are the lengths of the input and output sequences of training example $n$, respectively.

For the second method we model the transition probability with a neural network,
\begin{align}
\nonumber p(a_1 = i) &= \prod_{d=1}^{i-1} (1-p(e_{d,1}))p(e_{i,1}), \\
p(a_{j} = i | a_{j-1} = k) &= \prod_{d=k}^{i-1} (1-p(e_{d,j}))p(e_{i,j}),
\end{align}
where $p(e_{i,j})$ denotes the probability of \texttt{emit} for the alignment $a_{j}  = i$. This probability is obtained by feeding $[\mathbf{h}_{i};\boldsymbol{s}_j]$ into a feed forward neural network, 
\begin{equation}
p(e_{i,j}) = \sigma(\text{MLP}(\mathbf{W}_t[\mathbf{h}_{i};\mathbf{s}_j] + b_t)).
\end{equation}
For simplicity, $p(a_j=i|a_{j-1}=k, \mathbf{s}_j, \mathbf{h}_k^i)$ is abbreviated as $p(a_{j} = i | a_{j-1} = k)$.

\section{Training and Decoding}
Since there are an exponential number of possible alignments, it is computationally intractable to explicitly calculate every $p(\boldsymbol{y}, \boldsymbol{a}|\boldsymbol{x})$ and then sum them to get the conditional probability $p(\boldsymbol{y}|\boldsymbol{x})$. We instead approach the problem using a dynamic-programming algorithm similar to the forward-backward algorithm for HMMs \cite{rabiner1989tutorial}.  

\subsection{Training}
For an input $\boldsymbol{x}$ and output $\boldsymbol{y}$, 
the forward variable $\alpha(i,j) = p(a_j=i, \boldsymbol{y}_1^j | \boldsymbol{x})$.
The value of $\alpha(i,j)$ is computed by summing over the probabilities of every path that could lead to this cell. Formally, $\alpha(i,j)$ is defined as follows:

For  $i \in [1, I]$:
\begin{equation}
\begin{split}
\alpha(i, 1) 
& = p(a_1 = i ) p(y_1 | \mathbf{h}_i, \mathbf{s}_1).
\end{split}
\end{equation}

For $j \in [2, J]$, $i \in [1, I]$:
\begin{align}
\alpha(i,j) = &\ p(y_j|\mathbf{h}_i, \mathbf{s}_j) \cdot \\
&\sum_{k=1}^{i}\alpha(k, j-1)p(a_j = i | a_{j-1} = k). \nonumber
\end{align}

The backward variables, defined as $\beta(i,j) = p(\boldsymbol{y}^J_{j+1}|a_j=i,\boldsymbol{y}_1^j, \boldsymbol{x})$, are computed as:

For $i \in [1, I]$:
\begin{equation}
\beta(i, J) = 1.
\end{equation}

For $j \in [1, J-1]$, $i \in [1, I]$:
\begin{align}
\beta(i,j) = &\sum_{k=i}^{I}p(a_{j+1} = k | a_j = i)\beta(k, j+1) \cdot \nonumber \\
  &\qquad p(y_{j+1}|\mathbf{h}_k, \mathbf{s}_{j+1}). 
\end{align}

During training we estimate the parameters by minimizing the negative log likelihood of the training set $S$:
\begin{equation}
\begin{split}
\mathcal{L}(\boldsymbol{\theta}) &= - \sum_{(\boldsymbol{x}, \boldsymbol{y}) \in S} \log p(\boldsymbol{y}|\boldsymbol{x}; \boldsymbol{\theta})\\
&= - \sum_{(\boldsymbol{x}, \boldsymbol{y}) \in S} \log \sum_{i=1}^I\alpha(i, J).\\ 
\end{split}
\end{equation}

Let $\boldsymbol{\theta_j}$ be the neural network parameters w.r.t. the model output at position $j$. 
The gradient is computed as:
\begin{equation}
  \frac {\partial \log p(\boldsymbol{y} | \boldsymbol{x}; \boldsymbol{\theta})} 
        {\partial \boldsymbol{\theta}} 
  = \sum_{j=1}^J \sum_{i=1}^I 
      \frac {\partial \log p(\boldsymbol{y} | \boldsymbol{x}; \boldsymbol{\theta})} 
            {\partial \alpha(i, j)}
      \frac {\partial \alpha(i, j)} {\partial \boldsymbol{\theta_j}}.
\end{equation}
The derivative w.r.t. the forward weights is
\begin{equation}
  \frac {\partial \log p(\boldsymbol{y} | \boldsymbol{x}; \boldsymbol{\theta})} 
        {\partial \alpha(i, j)} 
  = \frac {\beta(i, j)} {p(\boldsymbol{y} | \boldsymbol{x}; \boldsymbol{\theta})}.
\end{equation}  

The derivative of the forward weights w.r.t. the model parameters at position $j$ is
\begin{align} 
  &\frac{\partial \alpha(i, j)} {\partial \boldsymbol{\theta_j}} 
  = \frac {\partial p(y_j | \mathbf{h}_i, \mathbf{s}_j)} {\partial \boldsymbol{\theta_j}}
   \frac{\alpha(i, j)} {p(y_j | \mathbf{h}_i, \mathbf{s}_j)} \nonumber \\
  &+ p(y_j | \mathbf{h}_i, \mathbf{s}_j) \sum_{k=1}^i \alpha(j-1, k) 
    \frac {\partial}{\partial \boldsymbol{\theta_j}} p(a_j{=}i | a_{j-1}{=}k). 
\end{align}  


For the geometric distribution transition probability model $ \frac {\partial} {\partial \boldsymbol{\theta_j}} p(a_j = i | a_{j-1} = k) = 0$.

\subsection{Decoding}

\begin{algorithm}                    
	\caption{DP search algorithm}        
	\label{decode} 
	\begin{algorithmic}                    
		\State \textbf{Input: } source sentence $\boldsymbol{x}$
		\State \textbf{Output: } best output sentence $\boldsymbol{y*}$
		\State \textbf{Initialization: } $Q \in \mathbb{R}^{I \times J_\text{max}}$, bp $\in \mathbb{N}^{I \times J_\text{max}}$,  $W \in \mathbb{N}^{I \times J_\text{max}}$, $I_\text{end}$, $J_\text{end}$.
		\For{$i \in [1, I]$}
			\State $Q[i,1] \gets \max_{y \in \mathcal{V}}p(a_1 = i) $$p(y|\mathbf{h}_i, \mathbf{s}_{1})$
			\State $bp[i,1] \gets 0$
			\State $W[i,1] \gets \argmax_{y \in \mathcal{V}}p(a_1 = i)$$p(y|\mathbf{h}_i, \mathbf{s}_{1})$
		\EndFor
		\For{$j\in[2, J_\text{max}]$}
			\For{$i \in [1, I]$}
				\State $Q[i,j] \gets \max_{y \in \mathcal{V}, k \in [1, i]} Q[k,j-1] \cdot$
				\State $\ \ \ \ \ \ \ \ \ \ \ \ \ \ \ p(a_j = i|a_{j-1} = k)p(y|\mathbf{h}_i, \mathbf{s}_{j})$
				\State $bp[i,j] , W[i,j] \gets \argmax_{y \in \mathcal{V}, k \in [1, i]}  \cdot$
				\State $Q[k,j-1]p(a_j = i|a_{j-1} = k)p(y|\mathbf{h}_i, \mathbf{s}_{j})$
			\EndFor
			\State $I_\text{end} \gets \argmax_i Q[i, j]$
			\If{$W[I_\text{end}, j] = \text{EOS}$ }
			\State $J_\text{end} \gets j$
			\Break
			\EndIf
		\EndFor 
		\State
		\Return a sequence of words stored in $W$ by following backpointers starting from $(I_\text{end}, J_\text{end})$.
	\end{algorithmic}
\end{algorithm}

For decoding, we aim to find the best output sequence $\boldsymbol{y}^*$ for a given input sequence $\boldsymbol{x}$:
\begin{equation}
\boldsymbol{y^*} = \argmax_{\boldsymbol{y}} p(\boldsymbol{y}|\boldsymbol{x})
\end{equation} 
The search algorithm is based on dynamic programming \cite{tillmann1997dp}. The main idea is to create a path probability matrix $Q$, and fill each cell $Q[i,j]$ by recursively taking the most probable path that could lead to this cell. 
We present the greedy search algorithm in Algorithm \ref{decode}. 
We also implemented a beam search that tracks the $k$ best partial sequences at position $(i,j)$.
The notation bp refers to backpointers, $W$ stores words to be predicted, $\mathcal{V}$ denotes the output vocabulary, $J_{\text{max}}$ is the maximum length of the output sequences that the model is allowed to predict.

\section{Experiments}
We evaluate the effectiveness of our model on two representative natural language processing tasks, sentence compression and morphological inflection. 
The primary aim of this evaluation is to assess whether our proposed architecture is able to outperform the baseline encoder-decoder model by overcoming its encoding bottleneck. We further benchmark our results against an attention model in order to determine whether our alternative alignment strategy is able to provide similar benefits while processing the input online.

\subsection{Abstractive Sentence Summarisation}

Sentence summarisation is the task of generating a condensed version of a sentence while preserving its meaning. In abstractive sentence summarisation, summaries are generated from the given vocabulary without the constraint of copying words in the input sentence. \newcite{DBLP:conf/emnlp/RushCW15} compiled a data set for this task from the annotated Gigaword data set \cite{graff2003english,napoles2012annotated}, where sentence-summary pairs are obtained by pairing the headline of each article with its first sentence. \newcite{DBLP:conf/emnlp/RushCW15} use the splits of 3.8m/190k/381k for training, validation and testing. 
In previous work on this dataset, \newcite{DBLP:conf/emnlp/RushCW15} proposed an attention-based model with feed-forward neural networks, and \newcite{chopra} proposed an attention-based recurrent encoder-decoder, similar to one of our baselines.

Due to computational constraints we place the following restrictions on the training and validation set: 

\begin{enumerate}
	\item The maximum lengths for the input sentences and summaries are 50 and 25, respectively.
	\item For each sentence-summary pair, the product of the input and output lengths should be no greater than 500. 
\end{enumerate}
We use the filtered 172k pairs for validation and sample 1m pairs for training. While this training set is smaller than that used in previous work (and therefore our results cannot be compared directly against reported results), it serves our purpose for evaluating our algorithm against sequence to sequence and attention-based approaches under identical data conditions.
Following from previous work \cite{DBLP:conf/emnlp/RushCW15,chopra,gulcehre2016pointing}, we report results on a randomly sampled test set of 2000 sentence-summary pairs.  The quality of the generated summaries are evaluated by three versions of ROUGE for different match lengths, namely ROUGE-1 (unigrams), ROUGE-2 (bigrams), and ROUGE-L (longest-common substring).

\begin{table}[t]\centering
	\begin{tabular}{@{}lccc@{}}
		\toprule
		Model &  ROUGE-1 & ROUGE-2  & ROUGE-L \\
		\midrule
		Seq2seq 			& 25.16 & 9.09 & 23.06 \\
		Attention	  & 29.25 & 12.85 & 27.32 \\
		\midrule
		uniSSNT	           & 26.96 & 10.54 & 24.59 \\
		biSSNT &  27.05 & 10.62 & 24.64\\
		uniSSNT+ & 30.15 & 13.59 & 27.88 \\
		biSSNT+   & \bfseries{30.27} & \bfseries{13.68} & \bfseries{27.91} \\
		\bottomrule
	\end{tabular}
	\caption {ROUGE F1 scores on the sentence summarisation test set. Seq2seq refers to the vanilla encoder-decoder and attention denotes the attention-based model. SSNT denotes our model with alignment transition probability modelled as geometric distribution. SSNT+ refers to our model with transition probability modelled using neural networks. The prefixes uni- and bi- denote using unidirectional and bidirectional encoder LSTMs, respectively.}
	\label{test_rg}
\end{table}

\begin{table}[t]\centering
	\begin{tabular}{@{}lcc@{}}
		\toprule
		Model &  Configuration & Perplexity  \\
		\midrule
		\multirow{4}{*}{Seq2seq} 			& $H=128, L=1$ & 48.5  \\
			  & $H=256,L=1$ & 35.6 \\
		  & $H=256, L=2$ & 32.1 \\
		  & $H=256, L=3$ & 31.0 \\
		\midrule
		 \multirow{2}{*}{biSSNT+}          & $H=128, L=1$ & 26.7 \\
		    & $H=256, L=1$ & \bfseries{22.6} \\
		\bottomrule
	\end{tabular}
	\caption{Perplexity on the validation set with 172k sentence-summary pairs.}
	\label{perp}
\end{table}

For training, we use Adam \cite{DBLP:journals/corr/KingmaB14} for optimization, with an initial learning rate of 0.001. The mini-batch size is set to 32. The number of hidden units $H$ is set to 256 for both our model and the baseline models, and dropout of 0.2 is applied to the input  of LSTMs. All hyperparameters were optimised via grid search on the perplexity of the validation set. We use greedy decoding to generate summaries.

Table \ref{test_rg} displays the ROUGE-F1 scores of our models on the test set, together with baseline models, including the attention-based model.
Our models achieve significantly better results than the vanilla encoder-decoder  and outperform the attention-based model. The fact that SSNT+ performs better is in line with our expectations, as the neural network-parameterised alignment model is more expressive than that modelled by geometric distribution.

To make further comparison, we experimented with different sizes of hidden units and adding more layers to the baseline encoder-decoder. Table \ref{perp} lists the configurations of different models and their corresponding perplexities on the validation set. We can see that the vanilla encoder-decoder tends to get better results by adding more hidden units and stacking more layers. This is due to the limitation of compressing information into a fixed-size vector. It has to use larger vectors and deeper structure in order to memorize more information. By contrast, our model can do well with smaller networks. In fact, even with 1 layer and 128 hidden units, our model works much better than the vanilla encoder-decoder with 3 layers and 256 hidden units per layer.

\subsection{Morphological Inflection}
Morphological inflection generation is the task of predicting the inflected form of a given lexical item based on a morphological attribute. 
The transformation from a base form to an inflected form usually includes concatenating it with a prefix or a suffix and substituting some characters. For example, the inflected form of a German stem {\it abgang} is {\it abg\"{angen} } when the case is dative and the number is plural.

In our experiments, we use the same dataset as \newcite{faruqui2015morphological}. This dataset was originally created by \newcite{durrett2013supervised} from Wiktionary, containing  inflections for German nouns (de-N), German verbs (de-V),  Spanish verbs (es-V), Finnish noun and adjective (fi-NA), and Finnish verbs (fi-V). It was further expanded by \newcite{nicolai2015inflection} by adding Dutch verbs (nl-V) and French verbs (fr-V). 
The number of inflection types for each language ranges from 8 to 57. The number of base forms, i.e. the number of instances in each dataset, ranges from 2000 to 11200.
The predefined split is 200/200 for dev and test sets, and the rest of the data for training. 

Our model is trained separately for each type of inflection, the same setting as the factored model described in \newcite{faruqui2015morphological}. The model is trained to predict the character sequence of the inflected form given that of the stem. The output is evaluated by accuracies of string matching. For all the experiments on this task we use 128 hidden units for the LSTMs and apply dropout of 0.5 on the input and output of the LSTMs. We use Adam \cite{DBLP:journals/corr/KingmaB14} for optimisation with initial learning rate of 0.001. During decoding, beam search is employed with beam size of 30.

\begin{table}[t]\centering
	\begin{tabular}{@{}lr@{}}
		\toprule
		{\bfseries{Model}} &  \bfseries{Avg. accuracy}\\
		\midrule
		Seq2Seq	& 79.08 \\
		Seq2Seq w/ Attention	 & 95.64  \\
		Adapted-seq2seq (FTND16)	  & \bfseries{96.20}  \\
		\midrule
		uniSSNT+ & 87.85 \\
		biSSNT+	 &   95.32\\
		\bottomrule
	\end{tabular}
	\caption{Average accuracy over all the morphological inflection datasets. The baseline results for Seq2Seq variants are taken from \protect\cite{faruqui2015morphological}. }
	\label{avg_acc}
\end{table}

Table \ref{avg_acc} gives the average accuracy of the uniSSNT+, biSSNT+, vanilla encoder-decoder, and attention-based models. The model with the best previous average result --- denoted as adapted-seq2seq (FTND16) \cite{faruqui2015morphological} --- is also included for comparison.  Our biSSNT+ model outperforms the vanilla encoder-decoder by a large margin and almost matches the state-of-the-art result on this task. As mentioned earlier, a characteristic of these datasets is that the stems and their corresponding inflected forms mostly overlap. Compare to the vanilla encoder-decoder, our model is better at copying and finding correspondences between prefix, stem and suffix segments.

\begin{table}[t]\centering
\begin{tabular}{@{}lcccc@{}}
	\toprule
	 Dataset & DDN13 & NCK15 & FTND16 & biSSNT+ \\
	\midrule
	 de-N	& 88.31 & \bfseries{88.60}  & 88.12 & 87.50\\
	 de-V	 & 94.76 & 97.50  & \bfseries{97.72} & 92.11 \\
	 es-V	  & 99.61 & 99.80  & \bfseries{99.81} & 99.52\\
	 fi-NA	 & 92.14 & 93.00  & 95.44 & \bfseries{95.48}\\
	 fi-V     & 97.23 & 98.10 & 97.81 & \bfseries{98.10}\\
	 fr-V    & 98.80 & \bfseries{99.20}  & 98.82 & 98.65 \\
	 nl-V    & 90.50 & 96.10 & 96.71 & 95.90\\ 
	 \midrule
	 Avg.    & 94.47 & 96.04  & \bfseries{96.20} & 95.32 \\
	\bottomrule
\end{tabular}
  \caption{Comparison of the performance of our model (biSSNT+) against the previous state-of-the-art on each morphological inflection dataset.}
\label{sep_acc}
\end{table}

Table \ref{sep_acc} compares the results of biSSNT+ and previous models on each individual dataset. DDN13 and NCK15 denote the models of \newcite{durrett2013supervised} and \newcite{nicolai2015inflection}, respectively. Both models tackle the task by feature engineering. FTND16 \cite{faruqui2015morphological} adapted the vanilla encoder-decoder by feeding the $i$-th character of the encoded string as an extra input into the $i$-th position of the decoder. It can be considered as a special case of our model by forcing a fixed diagonal alignment between input and output sequences. 
Our model achieves comparable results to these models on all the datasets. Notably it outperforms other models on the Finnish noun and adjective, and verbs datasets, whose stems and inflected forms are the longest.

\section{Alignment Quality}

\begin{figure*}[!htb]
	\begin{minipage}{0.5\textwidth}
		\begin{subfigure}{\linewidth}
			\begin{tikzpicture}
			\def \xscale {0.6}
			\def \yscale {0.3}

			\foreach \c [count=\x from 0] in {
				{\texttt{<}/s\texttt{>}},{.},{},{,},{director},{managing},{new},{a},{appointed},{has},{,},{daily},{business},{us-based},{the},{of},{edition},{asian},{the},{,},{asia},{journal},{street},{wall},{the} } 
			{
				\node [left] at (-\xscale,\x*\yscale) {\scriptsize \c};	
			}

			\foreach \c [count=\x from 0] in {
				{wall},{street},{journal},{asia}, {names},{new}, {managing},{director},{\texttt{<}/s\texttt{>}}}  
			{
				\node [rotate=90,right] at (\xscale*\x,25*\yscale) {\scriptsize \c};	
			} 
			
			\foreach \x in {0,...,8}
			\foreach \y in {0,1,3,4,...,24} 
			{
				\draw [white,fill=black!5]
				(\x*\xscale-0.5*\xscale,\y*\yscale-0.5*\yscale) rectangle (\x*\xscale+0.5*\xscale,\y*\yscale+0.5*\yscale);
			} 
			
			\node  at (-2*\xscale,2*\yscale) {\vdots};	
			\draw [fill]  
			(3.5*\xscale, 2*\yscale) circle [radius=0.04] 
			(4*\xscale, 2*\yscale) circle [radius=0.04]
			(4.5*\xscale, 2*\yscale) circle [radius=0.04]
			;	
			
			\draw [white,fill=blue!30] 
			(-0.5*\xscale,22.5*\yscale) rectangle (0.5*\xscale,23.5*\yscale)
			(0.5*\xscale,21.5*\yscale) rectangle (1.5*\xscale,22.5*\yscale)
			(1.5*\xscale,20.5*\yscale) rectangle (2.5*\xscale,21.5*\yscale)
			(2.5*\xscale,19.5*\yscale) rectangle (3.5*\xscale,20.5*\yscale)
			(3.5*\xscale,7.5*\yscale) rectangle (4.5*\xscale,8.5*\yscale)
			(4.5*\xscale,6.5*\yscale) rectangle (5.5*\xscale,7.5*\yscale)
			(5.5*\xscale,4.5*\yscale) rectangle (6.5*\xscale,5.5*\yscale)
			(6.5*\xscale,3.5*\yscale) rectangle (7.5*\xscale,4.5*\yscale)
			(7.5*\xscale,-0.5*\yscale) rectangle (8.5*\xscale,0.5*\yscale)
			;
			
			\end{tikzpicture}
			\caption{}
			\label{vis1}
		\end{subfigure}
	\end{minipage}    
	\begin{minipage}{.5\textwidth} 
		\begin{subfigure}{\linewidth}
			\begin{tikzpicture}
			\def \xscale {0.6}
			\def \yscale {0.3}

			\foreach \c [count=\x from 0] in {
				{\texttt{<}/s\texttt{>}},{n},{e},{k},{c},{o},{z}} 
			{
				\node [left] at (-\xscale,\x*\yscale) {\scriptsize \c};	
			}

			\foreach \c [count=\x from 0] in {
				{g},{e},{z},{o}, {c},{k}, {t},{\texttt{<}/s\texttt{>}}}  
			{
				\node [above] at (\xscale*\x,7*\yscale) {\scriptsize \c};	
			} 
			
			\foreach \x in {0,...,7}
			\foreach \y in {0,...,6} 
			{
				\draw [white,fill=black!5]
				(\x*\xscale-0.5*\xscale,\y*\yscale-0.5*\yscale) rectangle (\x*\xscale+0.5*\xscale,\y*\yscale+0.5*\yscale);
			}

			\draw [white,fill=blue!30] 
			(-0.5*\xscale,5.5*\yscale) rectangle (0.5*\xscale,6.5*\yscale)
			(0.5*\xscale,5.5*\yscale) rectangle (1.5*\xscale,6.5*\yscale)
			(1.5*\xscale,5.5*\yscale) rectangle (2.5*\xscale,6.5*\yscale)
			(2.5*\xscale,3.5*\yscale) rectangle (3.5*\xscale,4.5*\yscale)
			(3.5*\xscale,3.5*\yscale) rectangle (4.5*\xscale,4.5*\yscale)
			(4.5*\xscale,2.5*\yscale) rectangle (5.5*\xscale,3.5*\yscale)
			(5.5*\xscale,0.5*\yscale) rectangle (6.5*\xscale,1.5*\yscale)
			(6.5*\xscale,-0.5*\yscale) rectangle (7.5*\xscale,0.5*\yscale)
			;
			
			\end{tikzpicture}
			\caption{}
			\label{vis2}
		\end{subfigure}\\[1ex]
		\raggedleft
		\begin{subfigure}{\linewidth}
		\begin{tikzpicture}
			\def \xscale {0.5}
			\def \yscale {0.3}

			\foreach \c [count=\x from 0] in {
					{\texttt{<}/s\texttt{>}},{i},{t},{n},{y},{y},{m},{s},{u},{n}, {n},{e},{l},{a}} 
			{
					\node [left] at (-\xscale,\x*\yscale) {\scriptsize \c};	
				}

			\foreach \c [count=\x from 0] in {
					{a},{l},{e},{n}, {n},{u}, {s},{m},{y},{y},{n},{t},{i},{\texttt{<}/s\texttt{>}}}  
			{
					\node [above] at (\xscale*\x,14*\yscale) {\scriptsize \c};	
				} 
			
			\foreach \x in {0,...,13}
			\foreach \y in {0,...,13} 
			{
					\draw [white,fill=black!5]
					(\x*\xscale-0.5*\xscale,\y*\yscale-0.5*\yscale) rectangle (\x*\xscale+0.5*\xscale,\y*\yscale+0.5*\yscale);
				}

			\draw [white,fill=blue!30] 
			(-0.5*\xscale,12.5*\yscale) rectangle (0.5*\xscale,13.5*\yscale)
			(0.5*\xscale,11.5*\yscale) rectangle (1.5*\xscale,12.5*\yscale)
			(1.5*\xscale,9.5*\yscale) rectangle (2.5*\xscale,10.5*\yscale)
			(2.5*\xscale,9.5*\yscale) rectangle (3.5*\xscale,10.5*\yscale)
			(3.5*\xscale,8.5*\yscale) rectangle (4.5*\xscale,9.5*\yscale)
			(4.5*\xscale,6.5*\yscale) rectangle (5.5*\xscale,7.5*\yscale)
			(5.5*\xscale,6.5*\yscale) rectangle (6.5*\xscale,7.5*\yscale)
			(6.5*\xscale,5.5*\yscale) rectangle (7.5*\xscale,6.5*\yscale)
			(7.5*\xscale,4.5*\yscale) rectangle (8.5*\xscale,5.5*\yscale)
			(8.5*\xscale,3.5*\yscale) rectangle (9.5*\xscale,4.5*\yscale)
			(9.5*\xscale,2.5*\yscale) rectangle (10.5*\xscale,3.5*\yscale)
			(10.5*\xscale,1.5*\yscale) rectangle (11.5*\xscale,2.5*\yscale)
			(11.5*\xscale,0.5*\yscale) rectangle (12.5*\xscale,1.5*\yscale)
			(12.5*\xscale,-0.5*\yscale) rectangle (13.5*\xscale,0.5*\yscale)
			;
			
			\end{tikzpicture}
			\caption{}
			\label{vis3}
		\end{subfigure}
	\end{minipage}
	\caption{Example alignments found by BiSSNT+. Highlighted grid cells represent the correspondence between the input and output tokens.}
	\label{vis}
\end{figure*}
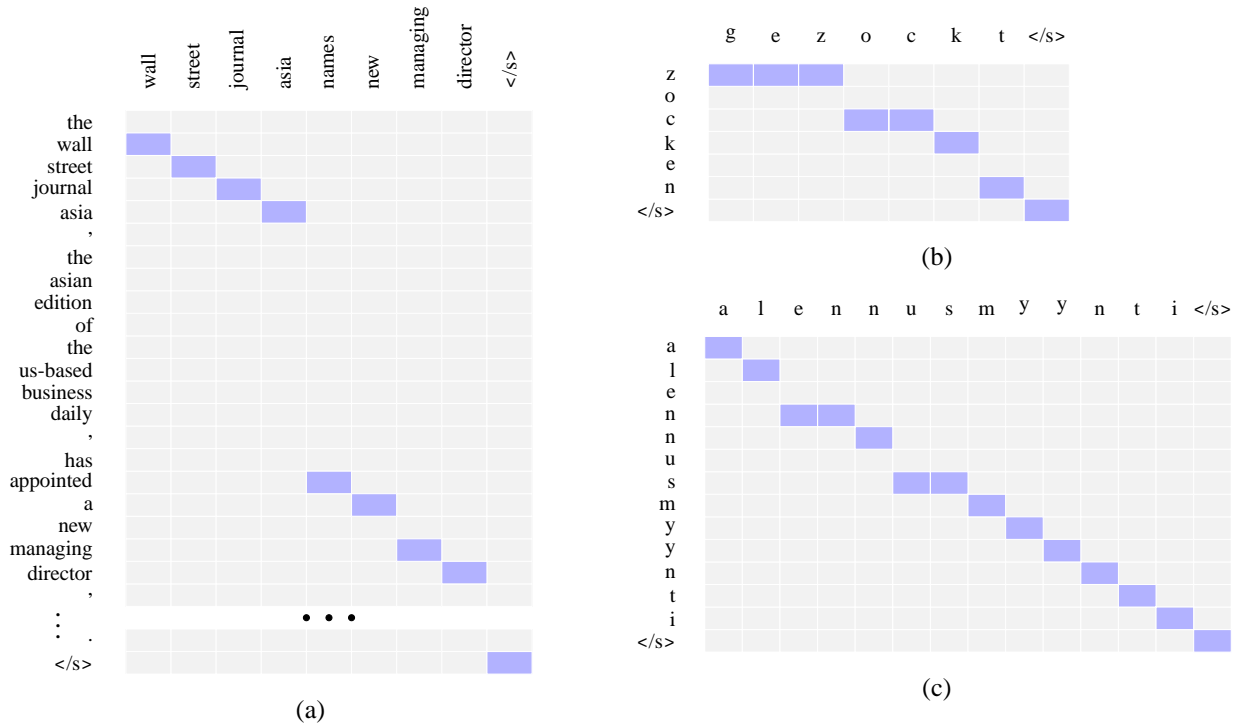

Figure \ref{vis} presents visualisations of segment alignments generated by our model for sample instances from both tasks. We see that the model is able to learn the correct correspondences between segments of the input and output sequences. For instance, the alignment follows a nearly diagonal path for the example in Figure \ref{vis3}, where the input and output sequences are identical. In Figure \ref{vis2}, it learns to add the prefix `ge' at the start of the sequence and replace `en' with `t' after copying `zock'. We observe that the model is robust on long phrasal mappings. As shown in Figure \ref{vis1}, the mapping between `the wall street journal asia, the asian edition of the us-based business daily' and `wall street journal asia' demonstrates that our model learns to ignore phrasal modifiers containing additional information. We also find some examples of word reordering, e.g., the phrase `industrial production in france' is reordered as `france industrial output' in the model's predicted output.

\section{Related Work}

Our work is inspired by the seminal HMM alignment model \cite{vogel1996hmm,tillmann1997dp} proposed for machine translation. In contrast to that work, when predicting a target word we additionally condition on all previously generated words, which is enabled by the recurrent neural models. This means that the model also functions as a conditional language model. It can therefore be applied directly, while traditional models have to be combined with a language model through a noisy channel in order to be effective. 
Additionally, instead of EM training on the most likely alignments at each iteration, our model is trained with direct gradient descent, marginalizing over all the alignments.

Latent variables have been employed in neural network-based models for sequence labelling tasks in the past. Examples include connectionist temporal classification (CTC) \cite{graves2006connectionist} for speech recognition and the more recent segmental recurrent neural networks (SRNNs) \cite{kong2015segmental}, with applications on handwriting recognition and part-of-speech tagging. Weighted finite-state transducers (WFSTs) have also been augmented to encode input sequences with bidirectional LSTMs~\cite{rastogi2016weighting}, permitting exact inference over all possible output strings.  
While these models have been shown to achieve appealing performance on different applications, they have common limitations in terms of modelling dependencies between labels. It is not possible for CTCs to model explicit dependencies. SRNNs and neural WFSTs model fixed-length dependencies, making it is difficult to carry out effective inference as the dependencies become longer. 

Our model shares the property of the sequence transduction model of \newcite{graves2012sequence} in being able to model unbounded dependencies between output tokens via an output RNN.
This property makes it possible to apply our model to tasks like summarisation and machine translation that require the tokens in the output sequence to be modelled highly dependently. 
\newcite{graves2012sequence} models the joint distribution over outputs and alignments by inserting null symbols (representing shift operations) into the output sequence. During training the model uses dynamic programming to marginalize over permutations of the null symbols, while beam search is employed during decoding.
In contrast our model defines a separate latent alignment variable, which adds flexibility to the way the alignment distribution can be defined (as a geometric distribution or parameterised by a neural network) and how the alignments can be constrained, without redefining the dynamic program. In addition to marginalizing during training, our decoding algorithm also makes use of dynamic programming, allowing us to use either no beam or small beam sizes.

Our work is also related to the attention-based models first introduced for machine translation~\cite{DBLP:journals/corr/BahdanauCB14}.
\newcite{DBLP:conf/emnlp/LuongPM15} proposed two alternative attention mechanisms: a global method that attends all words in the input sentence, and a local one that points to parts of the input words.  Another variation on this theme are pointer networks \cite{vinyals2015pointer}, where the outputs are pointers to elements of the variable-length input, predicted by the attention distribution.
\newcite{jaitly2015online} propose an online sequence to sequence model with attention that conditions on fixed-sized blocks of the input sequence and emits output tokens corresponding to each block. The model is trained with alignment information to generate supervised segmentations.

Although our model shares the same idea of joint training and aligning with the attention-based models, our design has fundamental differences and advantages.  While attention-based models treat the attention weights as output of a deterministic function (soft-alignment), in our model the attention weights correspond to a hidden variable, that can be marginalized out using dynamic programming. Further, our model's inherent online nature permits it the flexibility to use its capacity to chose how much input to encode before decoding each segment.

\section{Conclusion}
We have proposed a novel segment to segment neural transduction model that tackles the limitations of vanilla encoder-decoders that have to read and memorize an entire input sequence in a fixed-length context vector before producing any output. By introducing a latent segmentation that determines correspondences between tokens of the input and output sequences, our model learns to generate and align jointly. During training, the hidden alignment is marginalized out using dynamic programming, and during decoding the best alignment path is generated alongside the predicted output sequence. By employing a unidirectional LSTM as encoder, our model is capable of doing online generation.  Experiments on two representative natural language processing tasks, abstractive sentence summarisation and morphological inflection generation, showed that our model significantly outperforms encoder-decoder baselines while requiring much smaller hidden layers.  For future work we would like to incorporate attention-based models to our framework to enable such models to process data online.

\section*{Acknowledgments}
We thank Chris Dyer, Karl Moritz Hermann, Edward Grefenstette, Tom\'{a}\v{s}  K\v{o}cisk\'{y}, Gabor Melis, Yishu Miao and many others for their helpful comments. The first author is funded by EPSRC.

\bibliographystyle{emnlp2016}
\bibliography{emnlp2016}

\end{document}